\documentclass{article} % For LaTeX2e

\usepackage[preprint]{colm2026_conference}

\usepackage{microtype}
\usepackage{hyperref}
\usepackage{url}
\usepackage{lineno}
\usepackage[table]{xcolor}
\usepackage[normalem]{ulem}

\newcommand{\newval}[1]{\textcolor{black!50!black}{#1}}

\newcommand{\changed}[1]{\textcolor{black!50!black}{#1}}

\definecolor{darkblue}{rgb}{0, 0, 0.5}
\hypersetup{colorlinks=true, citecolor=darkblue, linkcolor=darkblue, urlcolor=darkblue}

\usepackage{algorithm}
\usepackage[noend]{algpseudocode}

\usepackage{graphicx}
\usepackage{bm}
\usepackage{amsthm, amssymb, amsmath}
\usepackage{xcolor,colortbl}
\usepackage{booktabs}
\usepackage{tabularx}
\usepackage{multirow}
\usepackage{array}
\usepackage[most]{tcolorbox}
\usepackage{soul}
\usepackage{caption}
\usepackage{cleveref}
\usepackage{wrapfig}

\newcolumntype{L}[1]{>{\raggedright\arraybackslash}p{#1}}
\newcolumntype{Y}{>{\raggedright\arraybackslash}X}

\theoremstyle{plain}

\theoremstyle{definition}

\theoremstyle{remark}

\definecolor{PTGreen}{RGB}{92,134,110}
\definecolor{PTGreenDark}{RGB}{52,92,75}
\definecolor{PTLightGreen}{RGB}{242,250,245}
\definecolor{PTWhite}{RGB}{255,255,255}
\definecolor{Gray}{gray}{0.9}

\newcommand{\V}[1]{\colorbox{black!7}{\texttt{\textbf{#1}}}} % highlighted variable

\tcbuselibrary{skins,breakable,raster}

\newcommand{\gC}{\mathcal{C}}

\crefname{figure}{Figure}{Figures}

\title{Semantic Refinement with LLMs for Graph Representations}

\author{
Safal Thapaliya$^{1,*}$, Zehong Wang$^{2,*}$, Jiazheng Li$^{1}$, Ziming Li$^{1}$,\\ \textbf{Yanfang Ye$^{2}$, Chuxu Zhang$^{1,\dag}$} \\
$^{1}$University of Connecticut, USA \quad $^{2}$University of Notre Dame, USA \\
\texttt{\{safal.thapaliya,jiazheng.li,ziming.li,chuxu.zhang\}@uconn.edu} \\
\texttt{\{zwang43,yye7\}@nd.edu} \\
{\small $^{*}$Equal Contribution \quad $^{\dag}$Corresponding Author}
}

\begin{document}

\ifcolmsubmission
\linenumbers
\fi

\maketitle

\begin{abstract}

    Graph-structured data exhibit substantial heterogeneity in where their predictive signals originate: in some domains, node-level semantics dominate, while in others, structural patterns play a central role.
    This structure--semantics heterogeneity implies that no graph learning model with a fixed inductive bias can generalize optimally across diverse graph domains.
    However, most existing methods address this challenge from the model side by incrementally injecting new inductive biases, which remains fundamentally limited given the open-ended diversity of real-world graphs.
    In this work, we take a data-centric perspective and treat node semantics as a task-adaptive variable.
    \changed{We propose a \textbf{G}raph-\textbf{E}xemplar-guided \textbf{S}emantic Refinement (\textbf{GES}) framework for graph representation learning which -- unlike existing LLM-enhanced methods that generate node descriptions without graph context -- leverages structurally and semantically similar nodes from the graph itself to guide semantic refinement. Specifically, a GNN is first trained to produce predictive states, which along with structural and semantic similarity are used to retrieve in-graph exemplars that inform an LLM in refining node descriptions.}
    We evaluate our approach on both text-rich and text-free graphs.
    \changed{Results show consistent improvements on semantics-rich and structure-dominated graphs, demonstrating the effectiveness of data-centric semantic refinement} under structure--semantics heterogeneity. 
    % The code is available at \url{https://anonymous.4open.science/r/GNN-LLM-18CF/}.
\end{abstract}

\section{Introduction}

Graph-structured data \citep{wu2020comprehensive, zhang2020deep} are ubiquitous in the real world, arising in diverse domains such as citation networks, social platforms, molecular interaction systems, and transportation infrastructures.
Despite sharing the same graph abstraction, these domains differ fundamentally in where their predictive signals originate.
In citation networks, for example, each node represents a scientific document whose topical content and research focus are explicitly encoded in natural language.
Here, node-level semantics---captured by titles, abstracts, or full texts---often provide the primary discriminative signal for downstream tasks, while citation links mainly serve as a contextual scaffold that propagates and regularizes semantic information \citep{greenberg2009citation,zhao2015analysis,zhang2019shne}.
By contrast, in domains such as molecular graphs or transportation networks, semantic attributes are weak or even absent \citep{wu2018moleculenet}.
Instead, node identity and functionality are determined predominantly by structural roles and global topological patterns, such as motifs, connectivity configurations, and relative positional relationships \citep{chen2020can,zhang2024beyond,wang2025towards}.
These examples demonstrate that predictive signals in real-world graphs may be dominated by semantics, dominated by structure, or arise from their intricate interplay.

This observation leads to a fundamental and unavoidable consequence:
\emph{the balance between semantics and structure is inherently domain-dependent, rather than governed by a universal principle}.
As a result, no graph learning model with a fixed inductive bias can perform optimally across graph domains with drastically different structure--semantics regimes \citep{platonov2023characterizing}.

However, translating this observation into a practical learning system remains challenging.
For a new graph, the dominant source of predictive signal—whether driven by semantics, structure, or their interaction—is unknown a priori, yet both the model and the data representation must commit to specific inductive biases in advance.
Modern GNNs encode fixed architectural preferences once chosen, favoring, for example, locality \citep{velickovic2018graph}, long-range dependencies \citep{xu2018how,rampavsek2022recipe}, or substructure information \citep{wang2025beyond,wang2025generative}.
Meanwhile, node representations—whether feature vectors \citep{kipf2017semisupervised}, textual embeddings \citep{wang2024gft}, or structural descriptors \citep{perozzi2014deepwalk,grover2016node2vec}—are typically constructed in a predefined manner and kept fixed throughout training.
As a result, the learning system becomes implicitly specialized to a particular structure--semantics regime.
When this specialization is mismatched with the true signal distribution of the target graph, performance degrades systematically, and adaptation in practice is often reduced to empirical model and feature selection rather than a principled mechanism.

To balance semantics and structure, most existing methods approach this problem primarily from the model side.
One line of work adapts GNN architectures by redesigning message passing \citep{morris2019weisfeiler,zhang2019heterogeneous,fan2022heterogeneous}, incorporating adaptive aggregation \citep{ying2018hierarchical}, or injecting positional encodings \citep{murphy2019relational}, thereby embedding different inductive biases into the model.
Beyond architectural modifications, another line of work introduces external reasoning models \citep{chen2024exploring,wang2023can}, most notably large language models (LLMs) \citep{zhao2023survey,ye2025llms4all}, which process graph structures and node attributes in textual form.
In parallel, other methods rely on auxiliary models \citep{chen2024exploring} to generate additional semantic signals—such as synthetic attributes \citep{he2023harnessing}—that are subsequently consumed by a downstream GNN.
Despite their empirical success, these approaches fundamentally rely on incrementally injecting model-level inductive biases, which cannot guarantee universal adaptability across open-ended and structurally diverse graph domains.

\begin{figure}[t]
    \centering
    \includegraphics[width=1.0\linewidth]{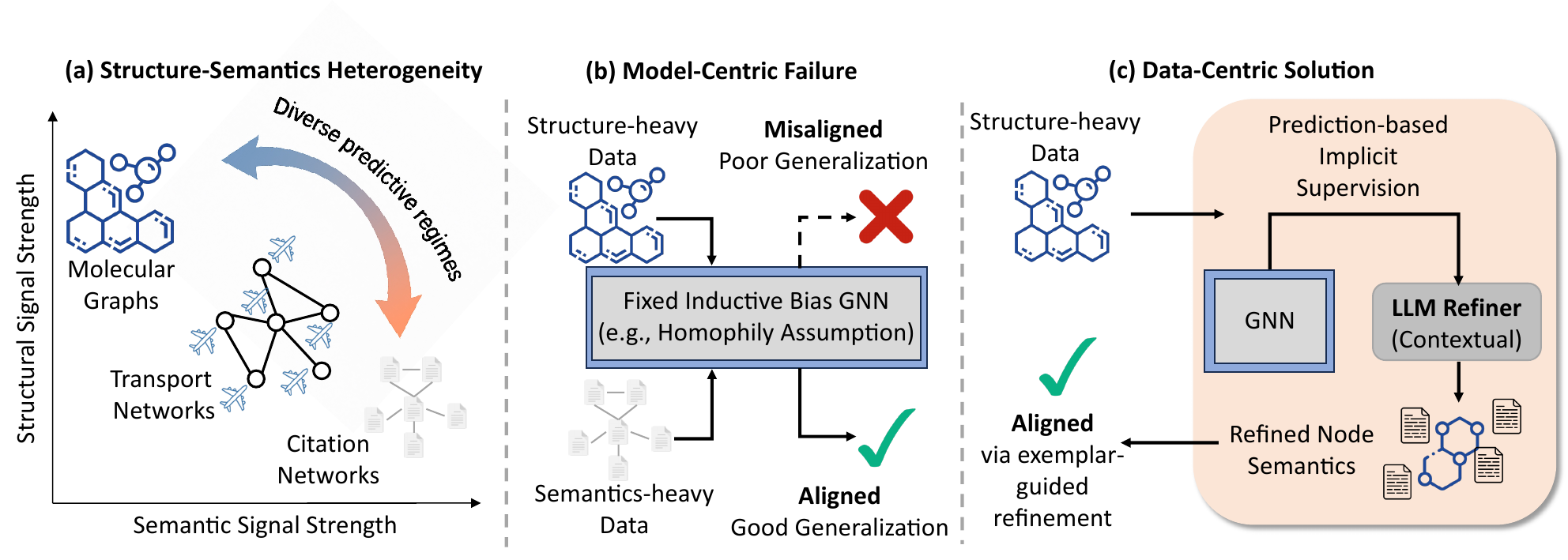}
  \vspace{-15pt}
  \caption{
    \textbf{Structure--semantics heterogeneity and data-centric adaptation.}
    (a) Real-world graphs vary widely in their reliance on semantic and structural patterns as sources of predictive signal.
    (b) Model-centric approaches with fixed inductive biases become misaligned when deployed across graphs with different structure--semantics regimes, leading to poor generalization.
    (c) In contrast, our proposed GES keeps the graph model fixed and refines node semantics through exemplar-guided refinement.
    }
    \vspace{-12pt}
    \label{fig:overview}
\end{figure}

In this work, we take a complementary data-centric perspective on structure-semantics heterogeneity by shifting the adaptation from the model to the data.
Instead of continually expanding model-level inductive biases, we treat node semantics as a task-adaptive variable.
This shift is motivated by the observation that the balance between structure and semantics is ultimately realized through the input representations consumed by the model, rather than through the architecture alone.
As a result, misalignment on new graph domains often arises from fixed node semantics that fail to reflect the graph-specific source of predictive signal.

Building on this perspective, we propose \textbf{G}raph-\textbf{E}xemplar-guided \textbf{S}emantic Refinement (\textbf{GES}), a data-centric framework for exemplar-guided semantic refinement.
Starting from initial node descriptions or structure-derived verbalizations \citep{wang2024tans}, we train a GNN for the downstream task and use its predictions as implicit supervision.
A large language model then refines node semantics by conditioning on both structural context and model behavior, and the refined descriptions are re-encoded for the final graph learner.
Through this single refinement pass, GES aligns node semantics with the structure--semantics regime of the target graph without modifying the underlying model.
% By iterating this closed loop for a small number of rounds, DAS progressively aligns node semantics with the structure--semantics regime of the target graph without modifying the underlying model.
We evaluate GES on both text-attributed and text-free graphs, where it consistently improves performance on structure-dominated and semantics-rich graphs.

\begin{figure*}[!ht]
  \centering
  \includegraphics[width=1.0\linewidth]{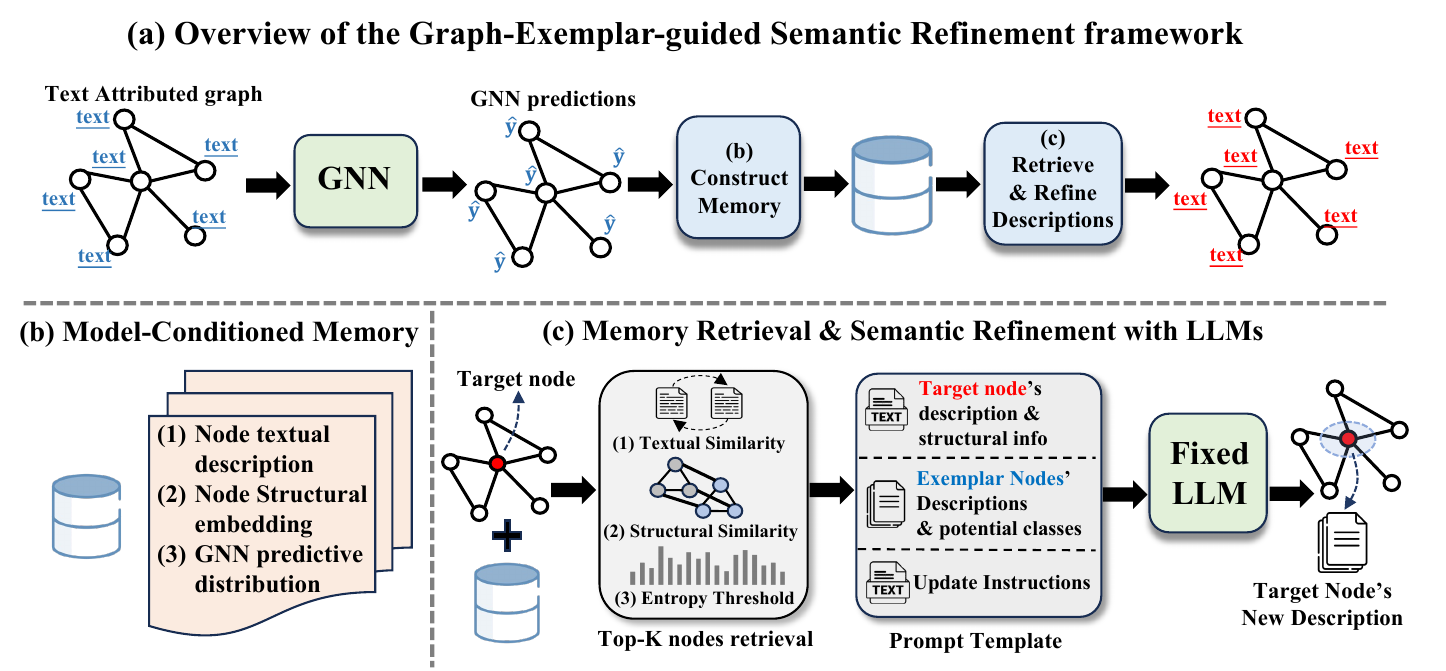}
  \vspace{-15pt}
  \caption{
    \textbf{Overview of the GES framework.}
    A GNN is first trained on initial node descriptions to populate a model-conditioned memory, from which structurally and semantically aligned in-graph exemplars are retrieved to guide an LLM in refining node semantics.
    % Node descriptions are iteratively refined through a closed loop between a fixed GNN and an LLM.
    % At each iteration, the GNN provides task feedback and a model-conditioned memory retrieves structurally and semantically aligned in-graph exemplars, which guide the LLM to update node semantics before feeding them back to the same GNN.
  }
  \label{fig:framework}
  \vspace{-12pt}
\end{figure*}

\section{Methodology}
\label{sec:method}

% We now introduce the proposed DAS framework.
% Our ultimate goal is to adapt node semantics under structural context and task feedback, so that a fixed graph neural network can operate effectively across different structure--semantics regimes.

\subsection{Problem Definition}

We consider a graph $G=(\mathcal{V}, \mathcal{E})$ with node set $\mathcal{V}$ and edge set $\mathcal{E}$.
Each node $v \in \mathcal{V}$ is associated with an initial description $r_v$, which is either a natural language text (in text-attributed graphs) or a structure-derived verbalization (in text-free graphs).
A subset of nodes $\mathcal{V}_{\mathrm{train}} \subseteq \mathcal{V}$ is labeled with $y_v \in \mathcal{Y}$.
Given a GNN $g_{\theta}$ for node classification, we treat node semantics as adaptive variables.
Our objective is to refine node descriptions $\{d_v\}_{v \in \mathcal{V}}$ from the initial inputs $\{r_v\}$, such that the refined semantics better align with both the graph structure and the downstream prediction task.
The final classifier $g_{\theta}$ trained on the refined descriptions $\{d_v'\}$ is used for evaluation.

\subsection{Overview}

GES is a data-centric framework that adapts node semantics under structural context and task supervision.
Instead of modifying model architectures to handle heterogeneous structure--semantics regimes, GES treats node descriptions as adaptive states that are refined through exemplar-guided LLM rewriting conditioned on GNN feedback.

% As illustrated in \cref{fig:framework}, DAS operates over $T$ refinement iterations.
% At iteration $t$, each node $v \in \mathcal{V}$ is associated with a description $d_v^{(t)}$ (initialized as $d_v^{(0)} = r_v$), which is encoded into node features and fed into a fixed GNN $g_{\theta}$ to produce predictions
As illustrated in \cref{fig:framework}, each node $v \in \mathcal{V}$ starts with an initial description $d_v$ (set to $r_v$), which is encoded into node features and fed into a GNN $g_{\theta}$ to produce predictions 
$\mathbf{p}_v = g_{\theta}(\mathbf{x}_v, G)$.
The descriptions and predictions are stored in a model-conditioned memory $\mathcal{B}$, from which an in-graph support set $\mathcal{S}_v$ is retrieved for each node.
Conditioned on $d_v$ and $\mathcal{S}_v$, the LLM refines the node semantics as
$d_v' = \mathcal{M}(d_v, \mathcal{S}_v)$.
The refined descriptions are then re-encoded and used to train the final GNN for evaluation.
% Through this feedback-driven loop, DAS progressively aligns node semantics with the structure-semantics regime of the target graph.
Unlike prior LLM-enhanced methods that rely on fixed prompts or exemplars \citep{he2023harnessing,chen2024exploring,wang2024tans}, GES enables task-conditioned semantic refinement guided by model behavior.

\subsection{Structure-Aware Initial Node Semantics}

We construct initial node descriptions by expressing structural information in natural language, so that both semantic and structural cues can be processed in a unified textual space.
Following \citet{wang2024tans}, for each node $v \in \mathcal{V}$ we compute a small set of structural statistics, including degree, betweenness, closeness, clustering coefficient, and square clustering coefficient \citep{zhang2017degree, saramaki2007generalizations, zhang2008clustering}. We present a detailed discussion in Appendix \ref{sec:graph-property}.

To eliminate scale variation across graphs, each statistic is converted into a percentile rank within the graph.
These normalized values are then mapped into a concise structural summary $t_v^{\text{struct}}$ via a fixed template, as shown in Appendix \ref{sec:prompt}.
For text-attributed graphs, we set $d_v = [r_v \,\Vert\, t_v^{\text{struct}}]$, while for text-free graphs we use $d_v = t_v^{\text{struct}}$.
This design expresses both semantic and structural information in a single textual modality, enabling consistent encoding and subsequent refinement.

\subsection{Model-Conditioned Memory}
\label{sec:buffer}

We maintain a model-conditioned memory to explicitly represent how node semantics, graph structure, and task predictions interact.
The memory is denoted as $\mathcal{B} = \{\beta_{v}\}_{v \in \mathcal{V}}$, which stores node-level states induced by the initial descriptions $\{d_v\}$ under the GNN $g_\theta$.
For each node $v \in \mathcal{V}$, the memory stores a joint state triple: $\beta_{v} = (d_v,\, \mathbf{s}_v,\, \mathbf{p}_v)$, where $d_v$ is the current textual description, $\mathbf{s}_v$ denotes a structure-oriented embedding encoding the node’s topological role via struc2vec \citep{ribeiro2017struc2vec}, and $\mathbf{p}_v$ is the predictive distribution produced by the GNN.
This triple defines a semantic--structural--predictive state for each node.

\paragraph{Memory Construction}
The memory $\mathcal{B}$ is constructed from the initial descriptions $\{d_v\}$, the fixed structural embeddings $\{\mathbf{s}_v\}$, and the GNN predictions obtained after training on these features.
This construction ensures that the memory reflects the alignment between node semantics, graph structure, and task-specific behavior under the current representation.

\paragraph{Memory Retrieval.}

Given the memory $\mathcal{B}$, the goal of memory retrieval is to identify, for each target node $v$, a small set of in-graph exemplars $\mathcal{S}_v$ that are simultaneously \emph{semantically relevant}, \emph{structurally aligned}, and \emph{reliable under the current classifier}.
These exemplars serve as task-aware references for subsequent semantic refinement.

To this end, the memory induces a joint semantic--structural similarity space.
Let $\mathbf{t}_v$ denote the embedding of the description $d_v$ produced by the text encoder \citep{wang2020minilm}, and let $\mathbf{s}_v$ denote the structural embedding encoding the topological role of node $v$.
For any pair of nodes $(v,u)$, we define the semantic similarity
$\text{sim}_t(v,u) = (\mathbf{t}_v^\top \mathbf{t}_u)/(\|\mathbf{t}_v\| \, \|\mathbf{t}_u\|)$
and the structural similarity
$\text{sim}_s(v,u) = (\mathbf{s}_v^\top \mathbf{s}_u)/(\|\mathbf{s}_v\| \, \|\mathbf{s}_u\|)$.
These two components are combined into a joint similarity score
\begin{equation}
  S(v,u) = \alpha \, \text{sim}_t(v,u) + (1-\alpha)\, \text{sim}_s(v,u),
\end{equation}
where $\alpha \in [0,1]$ controls the trade-off between semantic and structural proximity.
This design allows the retriever to adapt to different graph regimes, emphasizing textual semantics in text-rich graphs and structural roles in topology-dominated graphs.

For each target node $v$, all candidate nodes $u \in \mathcal{V}\setminus\{v\}$ are ranked according to $S(v,u)$.
From the top-ranked candidates, we further incorporate model confidence stored in the predictive state $\mathbf{p}_u$ to filter unreliable references.
Specifically, nodes with low predictive entropy are preferred.
The resulting exemplar set $\mathcal{S}_v$ thus consists of in-graph references that are not only close to $v$ in the joint semantic--structural space, but also stable with respect to the current task model.

Formally, for each target node $v$, we first rank all candidate nodes $u \in \mathcal{V}\setminus\{v\}$ by the joint similarity score $S(v,u)$.
Let $\mathcal{C}_v$ denote the top-$K$ candidates under this ranking.
We then define a confidence score for each candidate node $u$ based on the predictive distribution $\mathbf{p}_u$, for example using the normalized entropy
$H(u) = - \sum_{c \in \mathcal{Y}} p_u(c)\,\log p_u(c)$.
The final exemplar set is selected as
\begin{equation}
  \mathcal{S}_v = \Big\{\, u \in \mathcal{C}_v \;\Big|\; H(u) \le \tau \,\Big\},
\end{equation}
where $\tau$ is the entropy threshold that upper-bounds the normalized entropy, retaining only confidently classified exemplars.
This ensures that selected exemplars are both similar to $v$ in the joint semantic--structural space and reliable under the current classifier.

% \paragraph{Functional Role of the Memory.}
% The model-conditioned history memory serves as a task-aware state space that connects representation learning with predictive feedback.
% Unlike static feature buffers or heuristic exemplar pools, $\mathcal{B}^{(t)}$ evolves together with the node descriptions and the classifier, enabling exemplar selection and subsequent semantic refinement to be grounded in \emph{graph-specific and task-specific} evidence.
% This mechanism constitutes the core data-centric interface through which predictive feedback is transformed into progressive semantic adaptation.

\subsection{Memory-Guided Semantic Refinement}

Given the memory $\mathcal{B}$, GES updates node semantics through an in-context refinement operator.
This operator defines how the description of each node is locally reshaped under task-aligned, in-graph references.

\paragraph{Semantic Refinement Operator.}
For each node $v \in \mathcal{V}$, an exemplar set $\mathcal{S}_v \subset \mathcal{B}$ is first retrieved based on joint semantic--structural similarity and model stability.
The large language model $\mathcal{M}$ is then applied as a conditional refinement operator $d_v' = \mathcal{M}(d_v, \mathcal{S}_v)$,
where $d_v'$ denotes the refined semantic description.

The LLM is instructed to perform \emph{semantic reweighting and compression} rather than knowledge expansion.
Specifically, it reconstructs $d_v$ by emphasizing discriminative cues implicitly indicated by the exemplar set $\mathcal{S}_v$, while remaining faithful to the existing content.
Since $\mathcal{S}_v$ is drawn from the same graph and filtered by the current classifier, the refinement is implicitly shaped by both structural context and task supervision.

\paragraph{Parallel Update.}
The refinement is applied to all nodes in parallel, yielding
$\mathcal{D}' = \{ d_v' \mid v \in \mathcal{V} \}$.
These refined descriptions are re-encoded as node features and a new GNN is trained on these encodings for evaluation.

% \paragraph{Iterative Semantic Evolution.}
% Starting from the initialization $d_v^{(0)} = r_v$, the refinement proceeds as
% \begin{equation}
%   d_v^{(0)} \rightarrow d_v^{(1)} \rightarrow \cdots \rightarrow d_v^{(T)}.
% \end{equation}
% Early iterations primarily reflect coarse lexical and structural cues, while later iterations progressively concentrate on task-discriminative semantics shaped by model feedback.
% After $T$ iterations, the final descriptions $\{ d_v^{(T)} \}$ are used for evaluation.

% \subsection{Why Iterative Refinement Works}
% To further explain why DAS benefits from iterative semantic refinement, we provide a theoretical analysis in Appendix~\ref{sec:theory}.
% Specifically, we formalize DAS as an alternating optimization process over model parameters and node semantics, where the model-conditioned memory induces a task-adaptive surrogate objective.
% We show that, under mild assumptions, each refinement iteration is guaranteed not to increase this global objective.
% This result explains why DAS can progressively improve node semantics without suffering from uncontrolled semantic drift, and clarifies that the refinement loop constitutes a principled optimization process rather than heuristic text rewriting.

\section{Experiments}

\subsection{Experimental Setup}
\paragraph{Datasets.}
We evaluate on five graphs following \citet{wang2024tans}: two text-attributed citation networks, \texttt{Cora} and \texttt{Pubmed}, and three text-free airport networks, \texttt{USA}, \texttt{Europe}, and \texttt{Brazil} (Statistics are given in Table~\ref{tab:data}).
For \texttt{Cora}/\texttt{Pubmed}, nodes are papers (title+abstract), edges are citations, and classes are research topics.
For airports, nodes are airports, edges are flight connections, and classes correspond to activity levels \citep{ribeiro2017struc2vec}.

\paragraph{Baselines.}
For text-attributed graphs, we compare Raw Feat. (bag-of-words/TF-IDF), Raw Text (use original text), TAPE \citep{he2024harnessing}, KEA\citep{chen2024exploring}, and TANS \citep{wang2024tans}.
For text-free graphs, we compare hand-crafted topology features—Node Degree, Eigenvector \citep{dwivedi2023benchmarking}, Random Walk \citep{dwivedi2022graph}, and TANS \cite{wang2024tans}.
In all text-attributed baselines, generated texts are appended to the original node text and encoded by the same sentence encoder for fairness, while for text-free graphs, the generated texts are used directly as node descriptions.

\paragraph{Evaluation Protocol.}
Unless otherwise specified, we focus on node classification with a GCN backbone \citep{kipf2017semisupervised}.
We also report results with GAT \citep{velickovic2018graph} and MLP in the text-attributed setting, following \citet{wang2024tans}.
For single-graph learning, we adopt the low-label / high-label splits: in the low-label regime, we use 20/30 nodes per class for train/valid on \texttt{Cora}/\texttt{Pubmed} (10/20 for \texttt{Brazil}); in the high-label regime, we use a 60/20/20 train/valid/test split.
All reported numbers are averages over 30 random seeds with mean~$\pm$~standard deviation, selecting models by the best validation accuracy.
For the text encoder, we adopt MiniLM \citep{wang2020minilm} for fair comparison with other methods, unless otherwise noted.

\begin{table*}[!htbp]
    \centering
    \small
    \setlength{\tabcolsep}{6pt}
    \resizebox{\linewidth}{!}{
        \begin{tabular}{l l cccc cc c}
            \toprule
                                                    &              & \multicolumn{3}{c}{\textbf{Low-Label}} & \multicolumn{3}{c}{\textbf{High-Label}}                                                                                                               \\ \cmidrule(lr){3-5} \cmidrule(lr){6-8}
                                                    & Method       & \textbf{GCN}                           & \textbf{GAT}                            & \textbf{MLP}          & \textbf{GCN}          & \textbf{GAT}          & \textbf{MLP}          & A.R.        \\ \midrule
            \multirow{6}{*}{\rotatebox{90}{Cora}}   & Raw Feat.    & $78.39 \pm 1.69$                       & $79.31 \pm 1.70$                        & $66.18 \pm 4.95$      & $83.10 \pm 1.69$      & $82.45 \pm 1.23$      & $64.56 \pm 1.95$      & $6.00$      \\
                                                    & Raw Text     & $79.19 \pm 1.63$                       & $80.09 \pm 1.57$                        & $70.55 \pm 1.40$      & $87.45 \pm 1.15$      & $85.72 \pm 1.47$      & $78.95 \pm 1.45$      & $4.83$      \\
                                                    & + TAPE       & $79.64 \pm 1.36$                       & $80.28 \pm 1.37$                        & $70.97 \pm 2.02$      & $87.69 \pm 1.34$      & $86.21 \pm 1.33$      & $80.07 \pm 1.72$      & $3.50$      \\
                                                    & + KEA        & $80.08 \pm 1.71$                       & $79.80 \pm 1.58$                        & $70.72 \pm 1.51$      & $87.94 \pm 1.28$      & $86.58 \pm 1.10$      & $79.90 \pm 1.83$      & $3.67$      \\
                                                    & + TANS       & $80.66 \pm 1.77$                       & $80.86 \pm 1.65$                        & $72.82 \pm 1.52$      & $88.88 \pm 1.21$      & $88.20 \pm 1.55$      & $81.44 \pm 1.42$      & $2.00$      \\
            \rowcolor{gray!15}                      & + GES (Ours) &\newval{\bm{$82.40 \pm 1.88$}} & \newval{\bm{$82.09 \pm 1.41$}}                   & \newval{\bm{$75.68 \pm 1.41$}} & \bm{$89.31 \pm 1.14$} & \bm{$88.78 \pm 1.17$} & \newval{\bm{$82.78 \pm 1.87$}} & \bm{$1.00$} \\
            \midrule
            \multirow{6}{*}{\rotatebox{90}{Pubmed}} & Raw Feat.    & $75.39 \pm 1.51$                       & $74.59 \pm 1.36$                        & $68.01 \pm 1.99$      & $84.10 \pm 0.55$      & $84.31 \pm 0.66$      & $80.56 \pm 0.30$      & $6.00$      \\
                                                    & Raw Text     & $76.97 \pm 1.95$                       & $75.50 \pm 2.03$                        & $70.78 \pm 2.00$      & $87.49 \pm 0.54$      & $87.20 \pm 0.51$      & $82.58 \pm 0.38$      & $4.33$      \\
                                                    & + TAPE       & $76.50 \pm 3.27$                       & $75.30 \pm 1.92$                        & $71.06 \pm 2.13$      & $88.21 \pm 0.62$      & $87.80 \pm 0.48$      & $83.98 \pm 0.59$      & $3.83$      \\
                                                    & + KEA        & $76.88 \pm 1.73$                  & $75.74 \pm 2.06$                        & $71.32 \pm 2.51$      & $88.10 \pm 0.49$      & $87.77 \pm 0.50$      & $85.33 \pm 0.41$      & $3.33$      \\
                                                    & + TANS       & $76.27 \pm 2.35$                       & $76.99 \pm 2.02$                        & $73.64 \pm 2.59$      & $89.16 \pm 0.39$      & $87.98 \pm 0.48$      & $88.84 \pm 0.43$      & $2.50$      \\
        \rowcolor{gray!15}                      & + GES (Ours) & \newval{\bm{$79.52 \pm 1.89$ }}                      & \newval{\bm{$79.09 \pm 1.50$}}                   & \newval{\bm{$76.51 \pm 2.71$}} & \newval{\bm{$90.01 \pm 0.52$}} & \newval{\bm{$88.42 \pm 0.52$}} & \newval{\bm{$90.01 \pm 0.52$}} & \bm{$1.00$} \\
            \bottomrule
        \end{tabular}}
    \vspace{-8pt}
    \caption{Experimental results on text-attributed graphs. Boldface indicates the best and A.R. is the average ranking.}
    \label{tab:cora-pubmed-results}
    \vspace{-10pt}
\end{table*}
\begin{table*}[!htbp]
       \centering
       \small
       \setlength{\tabcolsep}{3pt}
       \resizebox{\linewidth}{!}{
              \begin{tabular}{l ccc ccc c}
                     \toprule
                                                   & \multicolumn{3}{c}{\textbf{Low-Label}} & \multicolumn{3}{c}{\textbf{High-Label}} &                                                                                                              \\\cmidrule(lr){2-4} \cmidrule(lr){5-7}
                     Method                        & Europe                                 & USA                                     & Brazil                 & Europe                & USA                   & Brazil                & A.R.        \\ \midrule
                     Raw Feat. (One-Hot)           & $51.89 \pm 2.75$                       & $52.74 \pm 2.25$                        & $65.15 \pm 15.93$      & $54.61 \pm 5.91$      & $60.88 \pm 3.83$      & $49.88 \pm 11.50$     & $5.83$      \\
                     Node Degree                   & $54.69 \pm 3.35$                       & $59.93 \pm 2.21$                        & $71.82 \pm 12.28$      & $55.72 \pm 5.12$      & $64.36 \pm 3.18$      & $63.83 \pm 9.35$      & $3.83$      \\
                     Eigenvector                   & $55.80 \pm 2.47$                       & $57.72 \pm 2.19$                        & $62.42 \pm 13.83$      & $58.15 \pm 4.51$      & $63.66 \pm 2.88$      & $65.06 \pm 8.95$      & $3.83$      \\
                     Random Walk                   & $56.70 \pm 2.47$                       & $56.11 \pm 2.11$                        & $69.70 \pm 14.34$      & $55.71 \pm 4.01$      & $62.80 \pm 3.01$      & $68.40 \pm 9.65$      & $4.00$      \\
                     TANS                          & $55.13 \pm 1.52$                       & $60.61 \pm 2.71$                        & {$80.61 \pm 12.14$} & $56.33 \pm 5.73$      & $65.81 \pm 3.11$      & $71.60 \pm 10.66$     & $2.50$      \\
                     \rowcolor{gray!15} GES (Ours) & $\bm{56.80 \pm 2.79}$                  & \newval{$\bm{61.66 \pm 1.77}$} & \newval{\bm{$80.91 \pm 11.10$}} & $\newval{\bm{59.51 \pm 4.44}}$& ${\bm{68.14 \pm 2.41}}$ & $\bm{75.19 \pm 7.42}$ & $\bm{1.00}$ \\
                     \bottomrule
              \end{tabular}
       }
       \vspace{-8pt}
       \caption{Experimental results on text-free graphs with GCN as backbone.}       \label{tab:airports_combined}
       \vspace{-10pt}
\end{table*}

\subsection{Main Results}
\subsubsection{Results on Text-Attributed Graphs}
Table~\ref{tab:cora-pubmed-results} reports node classification accuracy on text-attributed \texttt{Cora} and \texttt{Pubmed} under both low- and high-label settings, using GCN, GAT, and MLP backbones.
On these datasets, augmenting Raw Text with existing LLM-based methods consistently improves performance over Raw Feat.\ and Raw Text across all backbones in both label regimes.
GES further improves upon the other methods in all configurations, consistently achieving higher accuracy than TANS for all backbones.
These results indicate that exemplar-guided, structure-aware refinement is robust across architectures and scales of text-attributed citation graphs.

\begin{table*}[!t]
    \resizebox{\linewidth}{!}{
        \begin{tabular}{lccccccc c}
            \toprule
            \multicolumn{1}{r}{Source $\to$}    & \multicolumn{2}{c}{USA} & \multicolumn{2}{c}{Europe} & \multicolumn{2}{c}{Brazil} &                                                                                     \\ \cmidrule(lr){2-3} \cmidrule(lr){4-5} \cmidrule(lr){6-7}
            \multicolumn{1}{r}{Target $\to$}    & {Europe}                & {Brazil}                   & {USA}                      & {Brazil}              & {USA}                 & {Europe}              & A.R.        \\ \midrule
            Raw Feat. (One-Hot)                 & --                      & --                         & --                         & --                    & --                    & --                    & --          \\
            \quad + SVD                         & $30.55 \pm 4.61$        & $34.23 \pm 5.19$           & $45.90 \pm 3.90$           & $57.21 \pm 5.30$      & $24.95 \pm 3.19$      & $45.48 \pm 2.58$      & $5.00$      \\ \midrule
            Node Degree                         & $46.61 \pm 1.54$        & $52.29 \pm 3.91$           & $\mathbf{53.40 \pm 1.09}$  & $66.76 \pm 3.85$      & $54.35 \pm 2.22$      & $51.85 \pm 2.14$      & $2.17$      \\
            Eigenvector                         & $37.73 \pm 3.08$        & $32.79 \pm 4.49$           & $50.12 \pm 1.76$           & $61.49 \pm 4.33$      & $25.43 \pm 0.98$      & $50.96 \pm 4.42$      & $5.00$      \\
            Random Walk                         & $48.79 \pm 2.60$        & $58.13 \pm 3.38$           & $49.45 \pm 1.59$           & $62.38 \pm 5.98$      & $44.82 \pm 1.65$      & $52.71 \pm 2.09$      & $3.00$      \\
            TANS                                & $50.99 \pm 3.31$        & $67.17 \pm 4.68$           & $51.88 \pm 2.82$           & $71.59 \pm 3.97$      & \bm{$54.96 \pm 1.80$} & $53.79 \pm 2.15$      & $2.17$      \\
            \rowcolor{Gray} \textbf{GES (Ours)} & \bm{$51.96 \pm 3.04$}   & \bm{$68.78 \pm 3.80$}     & \textbf{$51.91 \pm 2.95$}  & \bm{$73.33 \pm 3.54$} & $54.20 \pm 1.99$      & \bm{$54.47 \pm 2.18$} & \bm{$1.67$} \\
            \bottomrule
        \end{tabular}
    }
    \vspace{-8pt}
    \caption{Experimental results on domain adaptation setting.}
    \label{tab:da}
    \vspace{-12pt}

\end{table*}

\vspace{-5pt}
\subsubsection{Results on Text-Free Graphs}
We evaluate GES on three text-free airport graphs using GCN as the backbone.
Table~\ref{tab:airports_combined} reports results under both low- and high-label splits.
In the low-label regime, structural baselines such as Node Degree, Eigenvector, and Random Walk outperform the one-hot feature baseline, confirming the importance of topology.
GES matches or surpasses these baselines on all three graphs, achieving the best accuracy across all datasets.
% In the high-label regime, GES yields clearer gains, achieving the strongest performance on \texttt{USA} and \texttt{Brazil}, and also outperforming all baselines on \texttt{Europe}.
Notably, although no human-written text is available, GES leverages structural cues to retrieve exemplars and induces task-aligned node semantics through exemplar-guided refinement.
% This structure-guided semantic refinement likely explains why gains on text-free graphs are larger than on text-attributed citation networks, where strong node texts already exist.
Overall, these results indicate that structure and semantic-aware refinement is particularly effective when raw node features are absent and structural roles dominate.

\vspace{-5pt}
\subsubsection{Results under Low Label Budget}
\begin{figure}[!h]
    \centering
    \includegraphics[width=\linewidth]{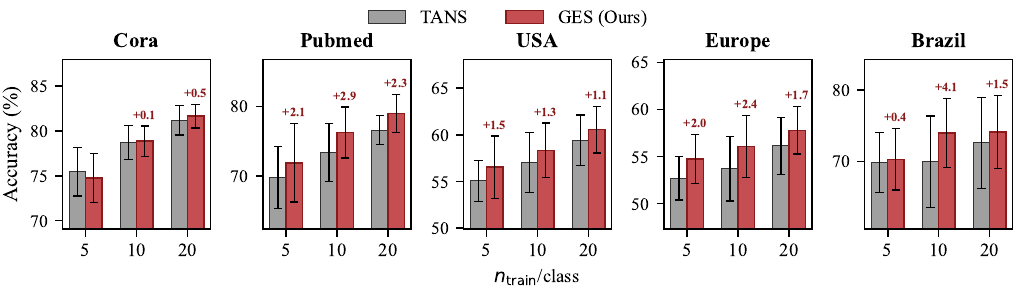}
    \vspace{-15pt}
    \caption{Label budget sensitivity across five datasets.}
    \label{fig:label-budget}
    \vspace{-12pt}
\end{figure}

We investigate how GES performs relative to TANS as the number of training labels per class varies.
Because GES descriptions are generated through exemplar retrieval from the model-conditioned memory, they encode implicit class structure even before the GNN is trained on the target labels.
When labeled data is abundant, the GNN has sufficient supervised signal to learn good representations regardless of description quality; when labels are scarce, the quality of input features becomes critical, and GES descriptions effectively provide soft supervision through the text itself.
Figure~\ref{fig:label-budget} confirms this: at 10 labels/class, GES outperforms TANS on all five datasets (avg.\ +2.16\%), with gains of +2.90\% on \texttt{Pubmed} and +4.06\% on \texttt{Brazil}.
At extreme scarcity (5 labels/class), GES wins on 4 of 5 datasets, with the sole exception being \texttt{Cora}, where rich text semantics already provide strong discriminative signal.
These results indicate that GES is most valuable in the realistic low-resource regime where labeled graph data is expensive to obtain.
\vspace{-12pt}

\vspace{-5pt}
\subsubsection{Results under Domain Adaptation}
\vspace{-3pt}

We assess cross-graph generalization in domain adaptation, where the model is trained on a source graph and directly evaluated on a target graph without fine-tuning.
We train on a source graph and evaluate on a target graph (20\% val / 80\% test on target).
We adopt the source-target pairs from \citet{wang2024tans} on the three airport graphs.
Results in Table~\ref{tab:da} show that TANS already improves over SVD-based alignment and purely structural features in most transfer directions.
GES further increases accuracy in the majority of source-target pairs.
For example, in \texttt{USA} to \texttt{Europe} or \texttt{Brazil}, and from \texttt{Europe} to \texttt{Brazil}, we achieve the best performance among all methods.
The performance gains of GES are especially notable on more challenging transfers involving \texttt{Brazil}, where structural roles differ markedly across graphs.
These improvements suggest that structure and semantic-aware node texts provide a more transferable representation than synthesized descriptions.

\subsubsection{Results under Transfer Learning}
\begin{wraptable}{r}{0.55\textwidth}
    \vspace{-12pt}
    \centering
    \small
    \setlength{\tabcolsep}{4pt}
    \begin{tabular}{lcc}
        \toprule
                                        & \texttt{C}$\to$\texttt{P} & \texttt{P}$\to$\texttt{C} \\ \midrule
        Raw Feat.\ + SVD               & $70.39 \pm 6.12$          & $70.48 \pm 3.71$          \\ \midrule
        Raw Text                        & $75.77 \pm 2.96$          & $79.62 \pm 2.04$          \\
        \quad + TAPE                    & $75.60 \pm 2.39$          & $79.25 \pm 2.06$          \\
        \quad + KEA                     & $75.25 \pm 2.50$          & $79.59 \pm 1.61$          \\
        \quad + TANS                    & $76.14 \pm 2.28$          & $80.05 \pm 1.74$          \\
        \rowcolor{Gray} \textbf{\quad + GES} & \bm{$79.80 \pm 1.76$} & \bm{$80.33 \pm 1.57$} \\
        \bottomrule
    \end{tabular}
        \vspace{-5pt}
    \caption{Pretrain--finetune transfer (GCN, low-label). C = Cora, P = Pubmed.}
    \label{tab:transfer}
    \vspace{-5pt}
\end{wraptable}

We also test whether GES produces node descriptions that generalize beyond the graph on which they were refined.
For this, we pretrain a GCN on a source citation graph and fine-tune on a target graph under the low-label split.
Table~\ref{tab:transfer} reports results for both directions between \texttt{Cora} and \texttt{Pubmed}.
GES achieves the best accuracy in both transfer directions, improving over TANS by +3.66\% on \texttt{Cora}$\to$\texttt{Pubmed} and +0.28\% on \texttt{Pubmed}$\to$\texttt{Cora}.
Notably, the \texttt{Cora}$\to$\texttt{Pubmed} gap is substantially larger than the single-graph improvement, suggesting that exemplar-guided descriptions capture class-discriminative cues that transfer more readily across related domains.
% This confirms that the semantic quality improvements of GES are not graph-specific artifacts but reflect genuinely more informative node representations.

\vspace{-5pt}
\section{Analysis}

\label{sec:mechanism-analysis}
This section analyzes the key mechanisms underlying GES beyond aggregate performance.
We focus on how exemplar retrieval and semantic refinement shape node representations. We provide additional analysis in Appendix \ref{sec:further-analysis} and \ref{sec:case-study}.

\vspace{-8pt}
% \subsection{The Role of Model-Conditioned Memory}
\paragraph{The Role of Model-Conditioned Memory.}
\begin{wraptable}{r}{0.5\textwidth}
    % \vspace{-12pt}
    \centering
    \small
    \begin{tabular}{lcc}
        \toprule
        \textbf{Retrieval} & \textbf{Cora} & \textbf{USA} \\
        \midrule
        GES (Ours) & \textbf{89.31 $\pm$ 1.14} & \textbf{68.14 $\pm$ 2.41} \\ \midrule
        Structure-only & 88.94 $\pm$ 1.07 & 66.29 $\pm$ 3.03 \\
        Text-only & 89.09 $\pm$ 1.01 & 66.38 $\pm$ 2.80 \\
        Random & 89.06 $\pm$ 0.92 & 66.47 $\pm$ 3.24 \\
        \bottomrule
    \end{tabular}
    \vspace{-5pt}
    \caption{Ablation on retrieval strategy.}
    \label{tab:ablation-memory}
    \vspace{-5pt}
\end{wraptable}
We examine the role of the model-conditioned memory in exemplar retrieval for semantic refinement.
To isolate its effect, we compare GES with three ablated variants: \emph{Random} exemplar selection, \emph{Structure-only} retrieval based solely on structural similarity, and \emph{Text-only} retrieval based solely on semantic similarity.
All variants use the same refinement procedure and prompt format.
% Results in Table~\ref{tab:ablation-memory} show that joint semantic--structural retrieval in GES consistently outperforms all ablated variants on both \texttt{Cora} and \texttt{USA}.
% On \texttt{Cora}, text-only retrieval performs competitively, reflecting the strong semantic signal in raw node texts, while structure-only retrieval is weaker.
% In contrast, on the text-free \texttt{USA} graph, both text-only and structure-only retrieval degrade performance, indicating that neither modality alone is sufficient for stable refinement.
% Random exemplar selection performs worst in both cases.
% These results demonstrate that effective refinement requires task-relevant exemplars that are aligned in both semantic content and structural role.
% The joint retrieval mechanism is particularly critical when node semantics must be induced from topology, where relying on a single modality can introduce noisy or misleading in-context signals.
Results in Table~\ref{tab:ablation-memory} show that joint semantic--structural retrieval in GES consistently outperforms all ablated variants on both \texttt{Cora} and \texttt{USA}.
On \texttt{Cora}, text-only retrieval performs competitively, reflecting the strong semantic signal in raw node texts, while structure-only retrieval is weaker.
In contrast, on the text-free \texttt{USA} graph, both single-modality variants degrade performance, indicating that neither modality alone is sufficient for stable refinement.

% \begin{figure}[!t]
%     \centering
%     \includegraphics[width=\linewidth]{images/embedding_discriminability.pdf}
%     \vspace{-15pt}
%     \caption{Embedding discriminability before GNN training. GES produces the most class-separable embeddings.}
%     \label{fig:discrim}
%     \vspace{-5pt}
% \end{figure}

% \begin{figure}[!t]
%     \centering
%     \includegraphics[width=0.65\linewidth]{images/retrieval_ablation.pdf}
%     \vspace{-5pt}
%     \caption{Retrieval strategy ablation (high-label, GCN).}
%     \label{fig:retrieval-ablation}
%     \vspace{-5pt}
% \end{figure}

\paragraph{Sensitivity to Retrieval Hyperparameters.}
\begin{wrapfigure}{r}{0.55\textwidth}
    \vspace{-12pt}
    \centering
    \includegraphics[width=0.55\textwidth]{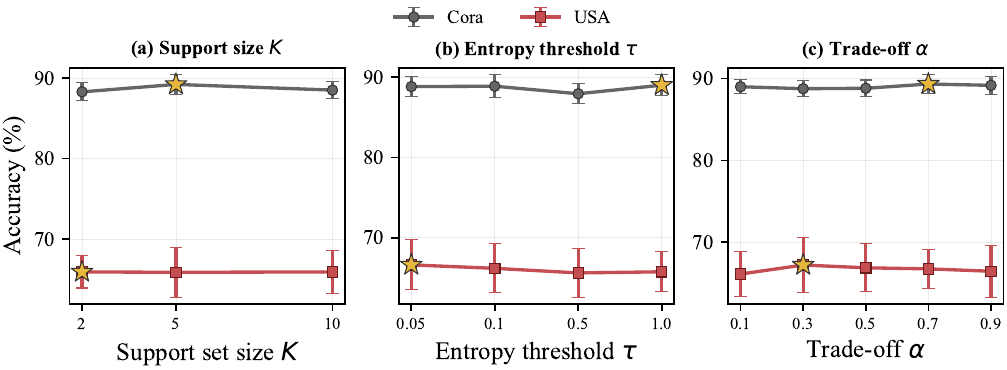}
    \vspace{-18pt}
    \caption{Sensitivity to $K$, $\tau$, and $\alpha$. Stars mark the best setting per dataset.}
    \label{fig:hyper-sensitivity}
    \vspace{-10pt}
\end{wrapfigure}

GES introduces three retrieval hyperparameters: the support-set size $K$, the entropy threshold $\tau$, and the semantic--structural trade-off $\alpha$ in $S(v,u)=\alpha\,\mathrm{sim}_t(v,u)+(1{-}\alpha)\,\mathrm{sim}_s(v,u)$.
Figure~\ref{fig:hyper-sensitivity} reports accuracy on \texttt{Cora} and \texttt{USA} (high-label, GCN) as each parameter is varied while the others are held at their defaults.
Performance is stable across a wide range of all three parameters.
Higher $\alpha$ and $\tau$ is preferred on text-rich \texttt{Cora}, while lower $\alpha$ and $\tau$ favors the text-free \texttt{USA}, consistent with the underlying structure--semantics regime.

\paragraph{Robustness to LLM Choice.}
\begin{wraptable}{r}{0.5\textwidth}
    \vspace{-12pt}
    \centering
    \small
    \setlength{\tabcolsep}{3pt}
    \begin{tabular}{l cc}
        \toprule
        \textbf{Dataset} & \textbf{TANS} & \textbf{GES} \\
        \midrule
        Cora   & $81.79 \pm 1.21$ & $\mathbf{82.00 \pm 1.51}$ \\
        Pubmed   & $83.00 \pm 1.98$ & $\mathbf{83.80 \pm 1.79}$ \\
        USA    & $60.61 \pm 3.23$ & $\mathbf{61.69 \pm 2.41}$ \\
        Europe & $56.40 \pm 3.64$ & $\mathbf{56.45 \pm 3.21}$ \\
        Brazil & $73.94 \pm 10.43$ & $\mathbf{75.45 \pm 13.92}$ \\
        \bottomrule
    \end{tabular}
    \vspace{-5pt}
    \caption{GES vs.\ TANS with Gemma-3-12B.}
    \label{tab:llm-ablation}
    \vspace{-5pt}
\end{wraptable}
Table~\ref{tab:llm-ablation} shows that replacing GPT-4o-mini with the open-weight Gemma-3-12B \citep{team2025gemma} in the low-label setting preserves the improvement of GES over TANS on all four datasets.
The gains are largest on structure-dominated graphs, where exemplar-guided refinement adds the most value regardless of LLM capacity.
This confirms that the improvements stem from the retrieval and refinement mechanism itself, not from the specific language model, and that GES can be deployed with open-weight models in resource-constrained settings.

\paragraph{Cost--Accuracy Tradeoff.}
\begin{wraptable}{r}{0.40\textwidth}
    \vspace{-12pt}
    \centering
    \small
    \setlength{\tabcolsep}{3pt}
    \begin{tabular}{l cc c}
        \toprule
        \textbf{Dataset} & \textbf{Extra tokens} & \textbf{$\Delta$Acc.} \\
        \midrule
        Cora     & +2.3M & +0.14 \\
        Pubmed  & +16.8M & +0.84 \\
        USA      & +1.0M & +2.33 \\
        Europe    & +0.3M & +3.18 \\
        Brazil   & +0.1M & +3.59 \\
        \bottomrule
    \end{tabular}
    \vspace{-8pt}
    \caption{Token overhead vs.\ accuracy gain of GES over TANS.}
    \label{tab:cost}
    \vspace{-10pt}
\end{wraptable}
A key practical question is whether the gains of GES justify its additional inference cost.
GES issues exactly one LLM call per node; the only overhead over other methods is the inclusion of retrieved exemplars in the prompt (${\sim}850$ additional input tokens per node).
Table~\ref{tab:cost} reports the token overhead of GES over TANS.
On \texttt{Pubmed}, the largest graph (19.7k nodes), this amounts to 16.8M extra tokens---substantial in absolute terms, but yielding +0.84\% in the high-label setting and +3.25\% in the low-label setting (\Cref{tab:cora-pubmed-results} and \Cref{tab:cost}).
On airport graphs, all three datasets combined require only 1.4M extra tokens, yet GES achieves +2.33\% on \texttt{USA}, +3.18\% on \texttt{Europe}, and +3.59\% on \texttt{Brazil}.
GES requires a single GNN pre-training pass to populate the memory, which is negligible relative to LLM inference.

% \begin{table}[!t]
%     \centering
%     \small
%     \setlength{\tabcolsep}{4pt}
%     \begin{tabular}{l ccc c}
%         \toprule
%         \textbf{Dataset} & \textbf{TANS cost} & \textbf{GES cost} & \textbf{Extra} & \textbf{$\Delta$ Acc.} \\
%         \midrule
%         Cora (2,708)     & \$0.47 & \$0.81 & +\$0.34 & +0.14 \\
%         Pubmed (19,717)  & \$3.40 & \$5.92 & +\$2.52 & +0.84 \\
%         USA (1,190)      & \$0.22 & \$0.48 & +\$0.26 & +2.33 \\
%         Europe (399)     & \$0.07 & \$0.16 & +\$0.09 & +3.18 \\
%         Brazil (131)     & \$0.02 & \$0.05 & +\$0.03 & +3.59 \\
%         \bottomrule
%     \end{tabular}
%     \caption{Cost--accuracy tradeoff (GPT-4o-mini). $\Delta$ Acc.: GES gain over TANS.}
%     \label{tab:cost}
% \end{table}

\section{Related Work}

We provide a more detailed discussion of related work in Appendix \ref{sec:comp-related-work}. 

\textbf{GNNs and Fixed Inductive Bias.}
Classical GNNs learn representations through local message passing, thereby encoding fixed inductive biases such as locality and homophily \citep{kipf2017semisupervised, hamilton2017inductive, velickovic2018graph, xu2018how}.
Structure-only methods based on random walks or structural roles further impose predefined topological assumptions \citep{perozzi2014deepwalk, grover2016node2vec, ribeiro2017struc2vec}.
While effective in specific regimes, these methods rely on fixed priors that do not adapt to heterogeneous structure--semantics distributions across graph domains.
\\
\textbf{Model-Centric Refinement.}
Most existing attempts to address structure--semantics heterogeneity remain \emph{model-centric}.
This includes strengthening GNN architectures with richer structural priors \citep{maron2018invariant, murphy2019relational, jin2022raw}, treating LLMs as direct graph reasoners via graph-to-text serialization \citep{zhao2023graphtext, kong2024gofa, chen2024llaga}, and introducing auxiliary models to enrich node representations with external semantic signals \citep{yao2019textgcn, he2023harnessing, yang2021graphformers}.
Despite their diversity, these approaches all inject additional inductive biases from the model side, and the construction of node representations remains largely static once the model is specified.
\\
\textbf{Data-Centric Refinement and Our Perspective.}
Existing data-centric methods primarily focus on graph augmentation, structure learning, or pseudo-labeling for robustness and generalization \citep{you2020graph, jin2020graph, chen2023label,he2023harnessing,chen2024exploring}, rather than on adapting node semantics to task-specific structure--semantics regimes.
In contrast, our work treats node semantics as task-adaptive variables and proposes a framework for exemplar-guided, structure-aware semantic refinement, providing a fundamentally different data-centric perspective for handling structure--semantics heterogeneity.

\vspace{-8pt}
\paragraph{The Position of Our Work.}
In contrast to the above data-centric paradigms, our work targets a fundamentally different objective.
Rather than augmenting data for invariance, modifying graph structure, or propagating pseudo-label supervision, we focus on \emph{task-driven refinement of node semantics themselves}.
Our method treats node semantics as \emph{adaptive variables} that are reshaped under structural context and predictive feedback from a downstream GNN, enabling direct handling of structure--semantics heterogeneity at the level where the balance between structure and semantics is instantiated.
\section{Conclusion and Limitations}

In this work, we proposed GES, a data-centric framework for exemplar-guided, structure-aware node semantic refinement on graphs.
By coupling a fixed GNN with a large language model through a model-conditioned memory, GES enables node semantics to be refined under joint structural context and task feedback.
Experiments on both text-rich and text-free graphs show that GES consistently improves over strong LLM-as-enhancer baselines in structure-dominated settings while remaining competitive in semantics-rich regimes.
More broadly, this work highlights a new direction for graph learning, where input representations are treated as dynamic, task-adaptive states rather than static features.

Several limitations remain.
First, GES requires a single LLM inference pass over all nodes, which incurs non-trivial overhead on large graphs; refining only uncertain or representative nodes offers a potential path toward scalability.
Second, the quality of refined descriptions depends on the underlying language model---weaker LLMs may fail to preserve structural cues, and even strong LLMs can occasionally introduce factual inconsistencies despite explicit prompt constraints.
Finally, our study focuses on node classification; whether the same refinement dynamics extend to link prediction, clustering, or graph-level tasks remains an open question for future work.

\bibliography{references}
\bibliographystyle{colm2026_conference}

\appendix

\section{Comprehensive Related Work}
\label{sec:comp-related-work}

\subsection{GNNs with Fixed Inductive Bias}

Classical graph representation learning is largely built upon fixed inductive biases that encode how structural and semantic information is propagated and aggregated over the graph.
Early message-passing GNNs \citep{kipf2017semisupervised,hamilton2017inductive,velickovic2018graph} propagate input node features through local neighborhood aggregation to learn task-specific representations.
By design, these architectures favor locality and homophily \citep{xu2018how,luan2022revisiting}, thereby imposing a strong but fixed prior on how predictive signals are assumed to distribute over the graph.
Complementary to message-passing models, structure-only methods characterize nodes by their positional or role similarity using random walks and structural homophily.
Representative approaches include DeepWalk \citep{perozzi2014deepwalk}, node2vec \citep{grover2016node2vec}, and struc2vec \citep{ribeiro2017struc2vec}, which learn embeddings purely from graph topology without relying on node semantics.
While highly effective in structure-dominated settings, these methods likewise rely on pre-specified structural assumptions that remain fixed across graph domains.
Motivated by this limitation, a large body of subsequent work has sought to address structure--semantics mismatch primarily from the model side, by designing new architectures and learning mechanisms with enhanced inductive biases.

\subsection{Model-Centric Adaptation}

\paragraph{Advanced GNN Model Architecture.}
A prominent line of model-centric approaches seeks to address structure--semantics heterogeneity by directly strengthening the inductive bias of GNN architectures \citep{xue2024self,han2024sadenas}.
These methods go beyond standard message passing by encoding richer structural priors into the model design.
For instance, \citet{maron2018invariant, chen2019equivalence, maron2019provably} propose $k$-order Weisfeiler--Lehman (WL) GNNs to emulate the $k$-WL test within neural architectures, while \citet{murphy2019relational, loukas2019graph} introduce positional and relational encodings to enhance representational power.
Another line of work leverages random walk kernels to guide the message passing process, further enriching the inductive bias of existing GNNs \citep{jin2022raw, tonshoff2021walking, wang2025beyond, wang2025generative}.
Despite improved expressivity, these models remain fundamentally model-centric: the inductive biases are still explicitly predefined by architecture design.
Moreover, many remain theoretically bounded by the $k$-WL hierarchy \citep{zhang2024beyond}, suggesting that architectural enhancement alone cannot offer a principled solution to the open-ended diversity of real-world graphs.

\paragraph{LLMs as Reasoners.}
Another emerging model-centric paradigm treats LLMs as direct graph reasoners.
These methods linearize graph structures and node attributes into natural language prompts and rely on the general reasoning capabilities of LLMs for training-free or lightly supervised graph classification and question answering \citep{zhao2023graphtext, guo2023gpt4graph, wang2023can, kong2024gofa, chen2024llaga}.
For example, \citet{wang2023can} describes graphs in natural language and applies LLMs to solve basic graph reasoning tasks, while GOFA \citep{kong2024gofa} and LLaGA \citep{chen2024llaga} operate over serialized graph representations or graph embeddings for downstream inference.
By bypassing explicit message passing, these approaches effectively replace graph-specific inductive biases with the intrinsic reasoning priors of LLMs.
However, this paradigm remains model-centric: structural information is processed solely according to the LLM and the serialization scheme, and context length limits together with the loss of explicit topology constrain scalability and long-range structural modeling.

\paragraph{Auxiliary Models to Enhance Graph Models.}
Beyond architectural modification and language-based reasoning, another class of model-centric approaches introduces auxiliary models to enrich the input representations of graph learners.
Early work such as TextGCN \citep{yao2019textgcn} demonstrates the benefit of incorporating external textual semantics into graph learning.
More recently, with the advent of LLMs, a growing body of methods leverage LLMs as semantic enhancers to generate or refine node descriptions for downstream GNNs \citep{chen2024text, zhang2024text, yan2023comprehensiveTAGreview, he2023harnessing, chen2024exploring, wang2024tans, yang2024latexgcl, fang2024gaugllm}.
In parallel, some works align auxiliary models with GNNs via joint training or embedding alignment \citep{yang2021graphformers, zhao2022learningglem, wen2023augmenting2p2}.
Despite their effectiveness, these approaches remain model-centric: auxiliary models inject additional semantic or structural inductive biases, while the resulting node representations are typically treated as static inputs by the downstream GNN rather than being refined under task-driven feedback.

\paragraph{Limitations of Model-Centric Methods.}
Despite their empirical success, the above paradigms share a fundamental commonality: they all address structure--semantics heterogeneity by injecting additional inductive biases from the model side.
Whether through architectural design, language-based reasoning, or auxiliary semantic enhancement, the manner in which semantic and structural information is combined is still determined by pre-specified model mechanisms.
However, real-world graph distributions are open-ended and structurally diverse, making it fundamentally impossible for any finite collection of model-level biases to guarantee universal adaptability.
Moreover, most model-centric approaches construct node representations in a largely static manner with respect to downstream learning dynamics, limiting their ability to adapt data representations to graph-specific structure--semantics regimes.

% \subsection{Data-Centric Adaptation}

% Apart from modifying the graph learning models, another line of works directly modify the graph data itself.
% Despite our methods can fall the category of the data-centric methods, the contribution of our works is to bring a new perspective to handle the semantics-structure heterogeneity problem, while existing data-centric works are not intended for this end.
% For existing works, they basically use data augmentation methods \citep{zhao2021data} like node feature corruption or graph structural corruption to let the model to learn the augmentation-invariant knowledge on graphs \citep{zhu2021graph,suresh2021adversarial,you2020graph,you2021graph,wang2023heterogeneous,wang2024select}.
% For example, GraphCL \citep{you2020graph} use four types of graph augmentation methods to generate different views of a single graph, enabling a single model is able to capture the discriminative information on graphs.
% Another line of works use graph structural learning to learn a better strucutre suitable for the learning of graph neural networks \citep{jin2020graph,liu2022towards,zhao2021heterogeneous,zhang2025mopi,perozzi2024let}.
% In addition, some of other works use pseudo-labels to guide the training of the model in limited label scenarios \citep{chen2023label}. However, the goal of our method is xxx.

\subsection{Data-Centric Adaptation}

Beyond modifying graph learning models, another line of research adopts a data-centric perspective by directly manipulating the graph data or input representations.
Most existing data-centric approaches are developed primarily for representation robustness, regularization, or generalization, rather than for explicitly addressing structure--semantics heterogeneity.

\paragraph{Graph Data Augmentation.}
A large body of work focuses on graph data augmentation, where node features or graph structures are perturbed to construct multiple views of the same graph for invariant representation learning \citep{zhao2021data, zhu2021graph, suresh2021adversarial, you2020graph, you2021graph, wang2023heterogeneous, wang2024select}.
For example, GraphCL \citep{you2020graph} applies a set of predefined structural and feature augmentations to generate contrastive graph views, enabling a model to capture augmentation-invariant information.
These methods are effective for improving robustness and transferability, but the underlying node semantics are not explicitly refined toward task-specific semantic--structural alignment.

\paragraph{Graph Structure Learning.}
Another line of data-centric work focuses on graph structure learning, which aims to optimize or reconstruct graph connectivity to better support GNN training \citep{jin2020graph, liu2022towards, zhao2021heterogeneous, zhang2025mopi, perozzi2024let}.
These approaches adapt the graph topology by removing spurious edges or adding task-relevant connections, thereby modifying the structural substrate on which message passing operates.
However, they primarily operate at the level of graph structure and do not directly model how node semantics should be adapted under different structure--semantics regimes.

\paragraph{Pseudo-Labeling.}
In addition, several studies explore pseudo-labeling and self-training schemes to guide representation learning in low-label settings \citep{chen2023label}.
While effective for label efficiency, such methods treat node features as fixed inputs and do not address the problem of task-driven semantic adaptation under structural context.

\section{Intuitive Motivation: Why Exemplar-Guided Refinement Produces Discriminative Embeddings}
\label{sec:theory}

We provide a theoretical motivation for why exemplar-guided semantic refinement produces more discriminative node representations. 
The key insight is that the model-conditioned memory induces a task-adaptive prototype structure, and the LLM refinement operator moves node embeddings toward class-coherent prototypes defined by reliable exemplars.

\subsection{Exemplar-Anchored Prototype Alignment}
Let $\mathcal{D}=\{d_v\}_{v\in\mathcal{V}}$ be the set of node descriptions and let $\mathbf{t}(d_v)$ denote the sentence embedding of $d_v$.
For each node $v$, the model-conditioned memory $\mathcal{B}$ provides an exemplar set $\mathcal{S}_v$ of structurally aligned, semantically similar, and confidently classified nodes.
Define the exemplar-induced prototype for node $v$ as
\begin{equation}
  \label{eq:anchor_def}
  \mathbf{m}_v
  =
  \frac{1}{|\mathcal{S}_v|}
  \sum_{u\in\mathcal{S}_v} \mathbf{t}(d_u).
\end{equation}
Because the exemplar set $\mathcal{S}_v$ is filtered by model confidence (low predictive entropy), nodes within $\mathcal{S}_v$ tend to be reliably classified.
When these exemplars share the same true class as $v$---which is encouraged by joint semantic--structural similarity---the prototype $\mathbf{m}_v$ approximates a class-coherent centroid in the embedding space.

The LLM refinement operator $d_v' = \mathcal{M}(d_v, \mathcal{S}_v)$ reconstructs the node description by emphasizing discriminative cues from the exemplar set.
In the embedding space, this operation can be interpreted as moving $\mathbf{t}(d_v')$ closer to the prototype $\mathbf{m}_v$, effectively reducing the within-class embedding variance while preserving between-class separation.
Formally, if the refinement satisfies
\begin{equation}
  \label{eq:refinement_contraction}
  \|\mathbf{t}(d_v') - \mathbf{m}_v\|_2 \le \|\mathbf{t}(d_v) - \mathbf{m}_v\|_2,
\end{equation}
then the global memory-consistency objective
\begin{equation}
  \label{eq:consistency_objective}
  \mathcal{R}(\mathcal{D}) = \sum_{v\in\mathcal{V}} \|\mathbf{t}(d_v) - \mathbf{m}_v\|_2^2
\end{equation}
is non-increasing after refinement.

\subsection{Connection to Embedding Discriminability}

The prototype alignment view predicts that GES refinement should increase the \emph{discriminability gap}---defined as the difference between average same-class and different-class cosine similarities in the embedding space.
We verify this prediction empirically in Section~\ref{sec:metric-validation}: GES descriptions exhibit a 17--36\% higher discriminability gap than TANS on citation graphs and a 26.4\% higher inter/intra cluster ratio on airport graphs.
Moreover, the discriminability gap is strongly correlated with downstream accuracy (Pearson $r=0.89$, $p<0.0001$), confirming that exemplar-guided refinement systematically improves the class-separability of node embeddings.

\section{Structural Features Used for Description Construction}
\label{sec:graph-property}

To characterize node-level structural roles in a compact yet informative manner, we employ a set of five widely used graph-theoretic measures. These features are chosen to balance descriptive power and computational efficiency, and are used solely to support structure-aware semantic construction.

\paragraph{Degree.}
The degree of a node $v$ is defined as the number of its immediate neighbors, reflecting its local connectivity within the graph:
\begin{equation}
  \deg(v) = |\mathcal{N}(v)|,
\end{equation}
where $\mathcal{N}(v)$ denotes the neighborhood of $v$. Nodes with higher degree typically correspond to locally influential or highly connected entities.

\paragraph{Betweenness Centrality.}
Betweenness centrality quantifies the extent to which a node lies on shortest paths between other node pairs, thereby capturing its bridging or mediating role in the network:
\begin{equation}
  \gC_B(v) = \sum_{s \neq v \neq t \in \mathcal{V}} \frac{\sigma_{st}(v)}{\sigma_{st}},
\end{equation}
where $\sigma_{st}$ is the total number of shortest paths between $s$ and $t$, and $\sigma_{st}(v)$ counts those paths that pass through $v$.

\paragraph{Closeness Centrality.}
Closeness centrality measures how close a node is, on average, to all other nodes in the graph. It is defined as
\begin{equation}
  \gC_C(v) = \frac{|\mathcal{V}|-1}{\sum_{u \in \mathcal{V}, u \neq v} d(u, v)},
\end{equation}
where $d(u,v)$ denotes the shortest-path distance between nodes $u$ and $v$. This measure reflects the global accessibility of a node.

\paragraph{Clustering Coefficient.}
The clustering coefficient evaluates the degree of local transitivity by measuring whether the neighbors of a node are also connected with each other:
\begin{equation}
  \gC_{\triangle}(v) = \frac{2T(v)}{\deg(v)(\deg(v)-1)},
\end{equation}
where $T(v)$ denotes the number of triangles that include node $v$. This metric captures the strength of tightly connected local neighborhoods.

\paragraph{Square Clustering Coefficient.}
Beyond triangular motifs, the square clustering coefficient characterizes the prevalence of four-node (quadrilateral) structures around a node. It reflects higher-order local dependencies and complementary structural patterns that are not captured by triangle-based clustering alone \citep{zhang2008clustering}.

% These structural quantities are later normalized and converted into textual summaries for use in structure-aware semantic refinement.

\section{Pseudo Code}

Algorithm~\ref{alg:GES} summarizes the overall GES pipeline.
GES first constructs a structure-aware textual summary for each node and (optionally) concatenates it with the raw node text to initialize descriptions.
It then trains a fixed GNN on the initial descriptions to populate a model-conditioned memory, retrieves in-graph exemplar sets for each node, and refines all node descriptions using an LLM conditioned on the retrieved exemplars.
The refined descriptions are used to train the final GNN classifier for evaluation.

\begin{algorithm}[t]
  \caption{GES Pipeline}
  \label{alg:GES}
  \small
  \begin{algorithmic}[1]
    \Require Graph $G=(\mathcal{V},\mathcal{E})$, optional node texts $\{r_v\}$, training labels on $\mathcal{V}_{\mathrm{train}}$, fixed-backbone GNN $g_\theta$, LLM $\mathcal{M}$, support size $K$
    \Ensure Final descriptions $\{d_v'\}$ and trained classifier $g_\theta^\ast$

    \State \textbf{Step 1: Preprocess Structural Information}
    \ForAll{$v \in \mathcal{V}$}
    \State Compute structural text $t_v^{\text{struct}}$
    \EndFor
    \State Compute structure-oriented embeddings $\{\mathbf{s}_v\}$

    \Statex

    \State \textbf{Step 2: Initialize Node Descriptions}
    \ForAll{$v \in \mathcal{V}$}
    \If{Graph is text-attributed}
    \State $d_v \gets [\, r_v \,\Vert\, t_v^{\text{struct}} \,]$
    \Else
    \State $d_v \gets t_v^{\text{struct}}$
    \EndIf
    \EndFor

    \Statex

    \State \textbf{Step 3: Construct Memory \& Refine Descriptions}
    \State Encode $\{d_v\}_{v\in\mathcal{V}}$ into features $\{\mathbf{x}_v\}$
    \State Train $g_\theta$ on labeled nodes $\mathcal{V}_{\mathrm{train}}$
    \State Obtain predictions $\{\mathbf{p}_v\}_{v\in\mathcal{V}}$
    \State Build memory $\mathcal{B} = \{(d_v, \mathbf{s}_v, \mathbf{p}_v)\}_{v\in\mathcal{V}}$
    \ForAll{$v \in \mathcal{V}$}
    \State Retrieve support set $\mathcal{S}_v \gets \textsc{Retrieve}(\mathcal{B}, K)$
    \State Refine description $d_v' \gets \mathcal{M}(d_v, \mathcal{S}_v)$
    \EndFor

    \Statex

    \State \textbf{Step 4: Final Training}
    \State Encode $\{d_v'\}$ into features and train final GNN $g_\theta^\ast$

    \Statex

    \State \Return $\{d_v'\}, g_\theta^\ast$
  \end{algorithmic}
\end{algorithm}

\section{Time Complexity}

Let $n = |\mathcal{V}|$ and $m = |\mathcal{E}|$. We summarize the cost of GES at a high level to clarify that the memory-based retrieval adds only modest overhead beyond the GNN and LLM calls. We assume bounded description/prompt lengths, fixed embedding dimensions, and fixed support size $K$.

\paragraph{One-time preprocessing.}
We compute structural statistics (for verbalized topology) and structure-oriented embeddings (e.g., struc2vec). This graph-dependent cost is incurred once:
\[
  C_{\text{pre}}(G) \;=\; C_{\text{stats}}(G) \;+\; C_{\text{struct-emb}}(G).
\]
\paragraph{Refinement cost.}
The single-pass refinement consists of three components:
\begin{align}
  C_{\text{refine}}(n,m)
   & = \underbrace{C_{\text{gnn}}(n,m) + n C_{\text{enc}} + \mathcal{O}(n)}_{\text{encode + GNN + memory construction}}
  \nonumber                                                                                                       \\
   & \quad+\underbrace{n C_{\text{sent}} + \mathcal{O}(n^2)}_{\text{exemplar retrieval}}
  \nonumber                                                                                                       \\
   & \quad+\underbrace{\mathcal{O}(nK) + n C_{\text{llm}}}_{\text{prompting + LLM refinement}} .
\end{align}
$C_{\text{enc}}$ is the description encoding cost, $C_{\text{sent}}$ is the sentence embedding cost used in retrieval, and $C_{\text{llm}}$ is the cost of one LLM call (dominated by prompt+generation tokens). The memory construction is only $\mathcal{O}(n)$; it does not introduce additional message passing or graph traversal beyond the fixed GNN backbone.

\paragraph{Total runtime.}
After refinement, we perform one final encoding and GNN training/evaluation on $\{d_v'\}$:
\[
  C_{\text{total}}
  = C_{\text{pre}}(G)
  + C_{\text{refine}}(n,m)
  + C_{\text{gnn}}(n,m) + n C_{\text{enc}}.
\]

\paragraph{Dominant terms.}
With bounded text length and fixed $K$, the main costs are (i) the GNN training/inference term $C_{\text{gnn}}(n,m)$, (ii) the retrieval term $\mathcal{O}(n^2)$ under brute-force similarity computation, and (iii) the LLM term $n C_{\text{llm}}$. All other components introduced by GES (memory writes, prompt assembly) are linear in $n$.

\section{Further Analysis}
\label{sec:further-analysis}

\subsection{Embedding Discriminability}
\label{sec:metric-validation}

\begin{figure}[!t]
    \centering
    \includegraphics[width=\linewidth]{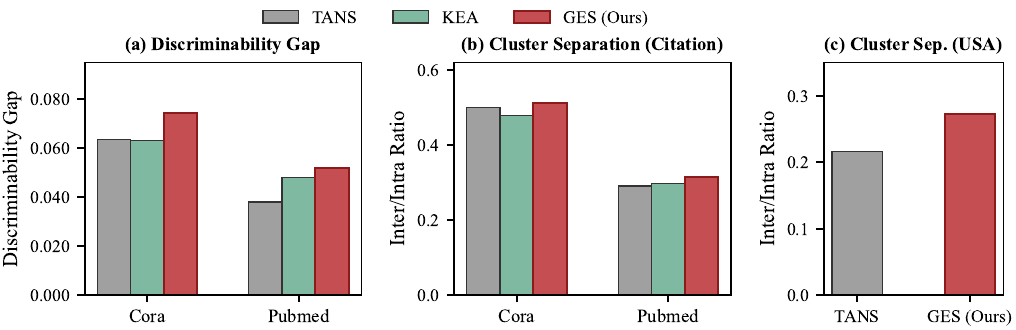}
    \caption{Embedding discriminability before GNN training. GES produces the most class-separable embeddings.}
    \label{fig:discrim}
\end{figure}

Because GES descriptions are conditioned on exemplars from structurally and semantically similar nodes, the LLM rewrites naturally emphasize class-consistent terminology.
We quantify this effect by measuring the \emph{discriminability gap}---the difference between average same-class and different-class cosine similarities in the MiniLM embedding space---and the \emph{inter/intra cluster ratio}, both computed before GNN training.
Figure~\ref{fig:discrim} reports these metrics across all methods.
GES achieves the highest discriminability gap on both \texttt{Cora} (+17.0\% over TANS) and \texttt{Pubmed} (+35.9\%), and the highest inter/intra ratio on the text-free \texttt{USA} graph (+26.4\%).
% Across all method--dataset combinations, the discriminability gap correlates strongly with downstream accuracy (Pearson $r{=}0.89$, $p{<}0.0001$; see Appendix~\ref{sec:metric-validation}), confirming that exemplar-guided refinement produces intrinsically more class-separable text representations, independent of the downstream classifier.

\paragraph{Metric validation.}
To confirm that the discriminability gap is a meaningful predictor of downstream performance, we compute correlations across all method--dataset combinations.
The Pearson correlation between discriminability gap and test accuracy is $r = 0.89$ ($p < 0.0001$), and the Spearman rank correlation is $\rho = 0.76$ ($p = 0.0006$).
Per-dataset, the correlation is near-perfect on \texttt{Pubmed} ($r = 0.99$, $p = 0.012$) and weak on \texttt{Cora} ($r = 0.05$, $p = 0.94$), consistent with the observation that text-rich graphs are less sensitive to description quality.
These results validate the discriminability gap as a reliable proxy for the quality of refined node descriptions.

\section{Case Study Results}
\label{sec:case-study}
\paragraph{Why GES Succeeds: Semantic Sharpening and Role Abstraction.}

GES is most effective when exemplar-guided refinement sharpens task-discriminative evidence already latent in the input representations.
On text-attributed graphs, successful refinement makes class-consistent technical cues more explicit, leading to reduced predictive entropy and corrections toward the true label.
On text-free graphs, GES succeeds when raw topological statistics are reorganized into a coherent semantic role interpretation (e.g., distinguishing regionally embedded nodes from global connectors based on clustering and betweenness).
In both cases, refinement aligns node semantics more closely with the structure--semantics regime of the graph, enabling the classifier to make more confident and accurate predictions. Representative examples are provided in Table \ref{tab:case-study-good}.

\paragraph{When GES Fails: Drift and Over-Confidence.}

Failure cases reveal inherent limitations of LLM-based semantic refinement.
One common failure mode is \emph{label drift}, where rewrites improve fluency without introducing additional discriminative evidence, causing predictions to shift toward a semantically adjacent but incorrect class.
Another failure mode is \emph{over-confidence}, in which refinement reduces predictive entropy while the prediction itself remains incorrect.
On text-free graphs, we additionally observe occasional \emph{attribute drift}, where numeric structural attributes are subtly altered during rewriting, raising faithfulness concerns even when predictive accuracy improves.
These failures highlight that effective refinement depends on maintaining a tight coupling between generated semantics and the underlying structural evidence.
See Table \ref{tab:case-study-bad} for examples.

\begin{table*}[t]
    \centering
    \small
    \setlength{\tabcolsep}{6pt}
    \renewcommand{\arraystretch}{1.15}
    \setlength{\emergencystretch}{2em}

    \begin{tabularx}{\textwidth}{L{0.28\textwidth} Y}
        \toprule
        \textbf{Dataset / node / labels} & \textbf{Node descriptions (original vs.\ refined)}                                                                                                                                                                                                                                                                                                                                                                                                                                                                                                                                                                                                                                                                                        \\
        \midrule

        \textbf{Dataset:} Cora\par
        \textbf{Node:} 1103\par
        \textbf{True label:} \textsc{Theory}\par
        \textbf{Pred:} \textsc{Theory}
                                         &
        \textbf{Original (default text):} Bias, variance and prediction error for classification rules. We...tain a bootstrap estimate of the error of a "bagged" classifier.\par\smallskip
        \textbf{Refined:} This paper delves into the relationship between bias, variance, and prediction error in classification frameworks, drawing on Efron's (1978) seminal work. It meticulously dissects prediction error into its fundamental components, elucidating their effects on classifier performance. The authors introduce a bootstrap-based estimate for the error associated with "bagged" classifiers, demonstrating how bagging and ensemble methods can significantly reduce prediction error. By providing a deeper theoretical understanding and practical tools, this work contributes to improving the accuracy and robustness of classification models.                                                                                           \\

        % \addlinespace[6pt]
        \midrule

        \textbf{Dataset:} Cora\par
        \textbf{Node:} 1148\par
        \textbf{True label:} \textsc{Theory}\par
        \textbf{Pred:} \textsc{Theory}
                                         &
        \textbf{Original (default text):} Error-based and entropy-based discretization of continuous... with respect to C4.5 and Naive-Bayesian classifiers on datas...\par\smallskip
        \textbf{Refined:} This paper presents a detailed comparison of error-based and entropy-based discretization techniques for machine learning classification tasks. It explores how discretization affects model performance and emphasizes the effectiveness of the MDL heuristic in guiding entropy-based discretization. The authors evaluate the methods using classifiers such as the C4.5 decision tree algorithm and Naive Bayes, demonstrating that the entropy-based MDL heuristic generally outperforms error minimization approaches. The study offers both theoretical insights and experimental validation, providing practical guidance for selecting discretization strategies that enhance classification accuracy.                                   \\

        % \addlinespace[6pt]
        \midrule

        \textbf{Dataset:} Europe-Airports\par
        \textbf{Node:} 333\par
        \textbf{True label:} \textsc{Medium-High Activity}\par
        \textbf{Pred:} \textsc{Medium-High Activity}
                                         &
        \textbf{Original (default text):} Given a node from a airport network graph, where the node ...etweenness Centrality" is 0.0000, ranked at 268 among 399 nodes.\par\smallskip
        \textbf{Refined:} This airport node is part of a network comprising 399 airports and displays graph characteristics that suggest medium-high activity. It has a Square Clustering Coefficient of 0.3393 (30th rank) and a Clustering Coefficient of 0.6364 (47th rank), indicating strong local clustering and well-connected neighboring airports. However, its degree is 8 (308th rank), showing it has relatively few direct connections compared to more central hubs. The node's Betweenness Centrality is 0.0000 (268th rank), suggesting it does not serve as a key connector between different parts of the network. Overall, this airport exhibits strong local connectivity but limited global influence, consistent with a medium-high activity airport. \\

        \bottomrule
    \end{tabularx}

    \caption{\textbf{Good refinement case studies (text-rich + text-free).}
        For each node, we show the \emph{original} (default\_text) description and the refined description,
        together with the ground-truth label and the predicted label.
        These examples illustrate successful refinement patterns: (i) in text-rich graphs (Cora), the rewrite
        adds or sharpens topic-discriminative technical cues consistent with the true class (e.g., theory/MDL/entropy-based
        discretization); (ii) in text-free airport graphs, the rewrite converts raw topological attributes into a
        coherent role/activity narrative that matches the correct activity bin.}
    \label{tab:case-study-good}
\end{table*}

\begin{table*}[t]
    \centering
    \small
    \setlength{\tabcolsep}{6pt}
    \renewcommand{\arraystretch}{1.15}
    \setlength{\emergencystretch}{2em}

    \begin{tabularx}{\textwidth}{L{0.28\textwidth} Y}
        \toprule
        \textbf{Dataset / node / labels} & \textbf{Node descriptions (original vs.\ refined)}                                                                                                                                                                                                                                                                                                                                                                                                                                                                                                                                                                                                                                                                                                                    \\
        \midrule

        \textbf{Dataset:} Cora\par
        \textbf{Node:} 392\par
        \textbf{True label:} \textsc{Genetic Algorithms}\par
        \textbf{Pred:} \textsc{Reinforcement Learning}
                                         &
        \textbf{Original (default text):} The Canonical Distortion Measure in Feature Space ...n. We establish some PAC-like sample complexity bounds ...tion problems. ...\par\smallskip
        \textbf{Refined:} This paper presents the Canonical Distortion Measure (CDM) as an innovative distortion metric designed for improved pattern comparison in machine learning tasks. It addresses shortcomings of traditional distortion measures and provides a theoretical framework with PAC-like bounds on sample complexity. The authors also validate CDM through experimental evaluation, demonstrating its practical relevance. Additionally, the study highlights CDM's applicability in feature space representation, offering enhanced performance in classification and pattern recognition problems. Overall, the work contributes to both theoretical understanding and practical implementation of distortion measures in machine learning.                                       \\

        \midrule
        % \addlinespace[6pt]

        \textbf{Dataset:} Cora\par
        \textbf{Node:} 714\par
        \textbf{True label:} \textsc{Genetic Algorithms}\par
        \textbf{Pred:} \textsc{Case Based}
                                         &
        \textbf{Original (default text):} Constructing Fuzzy Graphs from Examples: Methods t... locally independent fuzzy rules ...\par\smallskip
        \textbf{Refined:} This paper presents a novel and efficient algorithm for constructing fuzzy graphs from datasets by generating locally independent fuzzy rules. The method emphasizes computational efficiency and simplicity, enabling it to model uncertain or complex relationships in the data. It provides a user-friendly framework for representing structured knowledge through fuzzy graph structures, supporting applications in pattern recognition and knowledge representation. The algorithm offers a practical approach to capturing relationships and uncertainties inherent in real-world data.                                                                                                                                                                               \\

        % \addlinespace[6pt]
        \midrule

        \textbf{Dataset:} Europe-Airports\par
        \textbf{Node:} 322\par
        \textbf{True label:} \textsc{High Activity}\par
        \textbf{Pred} \textsc{Medium-Low Activity}
                                         &
        \textbf{Original (default text):} Given a node from a airport network graph, where the node ...ss Centrality" is 0.3909, ranked at 55 among 399 nodes.\par\smallskip
        \textbf{Refined:} This airport node is part of a network consisting of 399 airports and exhibits high local clustering, as shown by its clustering coefficient of 0.8125 (33rd rank) and square clustering coefficient of 0.4286 (77th rank). Despite this, its degree of 11 (260th rank) indicates relatively few direct connections, limiting its role as a major hub. Its closeness centrality rank of 55th suggests moderate overall reachability in the network. The node's betweenness centrality is relatively low compared to top connectors, implying it does not frequently lie on shortest paths between airports. Overall, these characteristics suggest that while the airport is well-integrated locally, it has limited global influence and fits a medium-low activity profile. \\

        \bottomrule
    \end{tabularx}

    \caption{\textbf{Failure case studies (text-rich + text-free).}
        We contrast nodes where refinement degrades performance or fails to correct an error.
        In text-rich graphs, failures arise when rewrites remain generic (improving fluency without adding class-specific
        evidence) or shift emphasis toward cues associated with an incorrect topic label.
        In text-free airport graphs, failures often come from over-interpreting or over-emphasizing a subset of structural
        signals (e.g., low degree / ``limited hub'' framing), which can push predictions toward an incorrect activity bin
        despite other metrics suggesting higher activity.}
    \label{tab:case-study-bad}
\end{table*}

\section{Dataset Statistics}
Table~\ref{tab:data} summarizes the five benchmark graphs used in our experiments.
\texttt{Cora} and \texttt{Pubmed} are text-attributed citation networks where each node carries a title and abstract; \texttt{USA}, \texttt{Europe}, and \texttt{Brazil} are text-free airport graphs where node identity is determined entirely by structural role.
The datasets span a wide range of scales (131--19,717 nodes) and structure--semantics regimes, providing a comprehensive testbed for evaluating data-centric semantic refinement.

\begin{table}[!h]
  \centering
  \small
    \begin{tabular}{lrrrl}
      \toprule
                      & \textbf{Nodes} & \textbf{Edges} & \textbf{Classes} & \textbf{Graph Types} \\ \midrule
      \texttt{Cora}   & 2{,}708        & 10{,}556       & 7                & Text-attribute       \\
      \texttt{Pubmed} & 19{,}717       & 88{,}648       & 3                & Text-attribute       \\ \midrule
      \texttt{USA}    & 1{,}190        & 28{,}388       & 4                & Text-free            \\
      \texttt{Europe} & 399            & 12{,}385       & 4                & Text-free            \\
      \texttt{Brazil} & 131            & 2{,}137        & 4                & Text-free            \\ \bottomrule
    \end{tabular}
  \caption{Dataset statistics used in our experiments.}
  \label{tab:data}
\end{table}

\section{Implementation Details}
\label{app:implementation}

% \paragraph{Backbones, encoders, and data splits.}
% We follow the experimental protocol described in the main text.  
% All methods (baselines and DAS) are evaluated on the five graphs Cora, Pubmed, USA, Europe, and Brazil using the single-graph low-/high-label splits from \citet{wang2024tans}.
% Unless otherwise specified, we use a GCN backbone for node classification; on text-attributed graphs we additionally report results with GAT and an MLP encoder.
% For node texts we adopt MiniLM \cite{wang2020minilm} as the sentence encoder, and for structure-oriented node representations we use \textsc{struc2vec} \citep{ribeiro2017struc2vec}.
% All reported numbers in the main tables are averages over 30 random seeds with mean and standard deviation, and model selection is based on validation accuracy.

% \paragraph{Hyperparameter search.}
For each dataset, backbone, and method we perform a random search over architecture and optimization hyperparameters.
The candidate values are:
hidden dimension \{8, 16, 32, 64, 128, 256\},
number of layers \{1, 2, 3\},
normalization layer in \{none, batchnorm\},
learning rate \{5e-2, 1e-2, 5e-3, 1e-3\},
weight decay \{0.0, 5e-5, 1e-4, 5e-4\},
and dropout rate \{0.0, 0.1, 0.5, 0.8\}.
For each configuration we choose the setting that achieves the best validation accuracy.
GES introduces two additional hyperparameters: the entropy threshold $\tau$ and the support-set size $K$ used in exemplar retrieval.
For citation graphs, We fix $K=10$ and $\tau = 0.5$ for all datasets.

\section{Prompts}
\label{sec:prompt}

\paragraph{Node initialization.}
For each node, we build an initial description by concatenating (i) an \textit{original description} and
(ii) a \textit{topological summary}.
For citation graphs, the original description is the node text (e.g., paper title and abstract).
For datasets without node text (e.g., airport graphs), this component is omitted.
The topological summary encodes structural cues using a fixed natural-language schema: we first state the
global graph context (graph type, node type, number of nodes, edge type, and number of edges), and then
append node-level property statements where each property is reported with its scalar value and its rank
among all nodes. The exact verbalization template is shown in Figure~\ref{tab:topological-template} and
follows the consistent schema adapted from \citet{wang2024tans}.

\paragraph{Semantic refinement.}
Starting from the initialized description above, we refine the target node text using a single unified prompt wrapper. 
Only the placeholders (highlighted tokens such as \V{GRAPH\_TYPE}) are swapped per node/dataset. 
An optional \textit{example} block provides a small set of training nodes as reference for how descriptions behave under the GNN, serving as in-context calibration data. 
Figure~\ref{fig:prompt-templates} shows the complete prompt layout.

\begin{figure*}[t]
  \centering
  \begin{tcolorbox}[
      title=\textbf{Topological Summary Template (used in initialization)},
      colframe=PTGreenDark,
      colback=PTLightGreen,
      colbacktitle=PTGreenDark,
      coltitle=white,
      fonttitle=\bfseries,
      boxrule=0.9pt,
      arc=0pt,
      left=7pt,right=7pt,top=7pt,bottom=7pt,
      width=\textwidth,
      enhanced
    ]
    \ttfamily\footnotesize
    Given a node from a \V{GRAPH\_TYPE} graph, where the node type is \V{NODE\_TYPE} with \V{NUM\_NODES} nodes,\\
    and the edge type is \V{EDGE\_TYPE} with \V{NUM\_EDGES} edges.\\[0.4ex]
    The value of property ``\V{PROPERTY\_NAME}'' is \V{PROPERTY\_VALUE}, ranked at \V{PROPERTY\_RANK} among \V{NUM\_NODES} nodes.\\
    (Repeat the property sentence for each selected topological feature.)
  \end{tcolorbox}

  \caption{Fixed verbalization template used to construct the topological summary in node initialization.}
  \label{tab:topological-template}
\end{figure*}

\begin{figure*}[t]
  \centering
  \begin{tcolorbox}[
      title=\textbf{Unified Refinement Prompt Template},
      colframe=PTGreenDark,
      colback=PTLightGreen,
      colbacktitle=PTGreenDark,
      coltitle=white,
      fonttitle=\bfseries,
      boxrule=0.9pt,
      arc=0pt,
      left=7pt,right=7pt,top=7pt,bottom=7pt,
      width=\textwidth,
      enhanced
    ]
    \ttfamily\footnotesize
    [System]\\
    You are rewriting node descriptions to make them clearer and more discriminative for a graph classifier.\\
    Each node is from a \V{GRAPH\_TYPE}.\\[0.6ex]

    [Target node]\\
    Original description: \V{ORIGINAL\_DESCRIPTION}\\
    Topological summary: \V{TOPOLOGICAL\_DESCRIPTION}\\[0.6ex]

    [Optional: Target Predictive Performance]\\
    GNN prediction = \V{PRED\_NAME} (\V{CORRECTNESS}); top prob = \V{TOP\_PROB}; entropy = \V{ENTROPY\_VAL}.\\
    Description used: \V{TEXT\_DESC}\\[0.6ex]

    [Optional: Training examples for reference]\\
    Example \V{ITEM\_NUMBER}:\\
    \hspace*{1.2em}Original: \V{EX\_ORIGINAL\_DESCRIPTION}\\
    \hspace*{1.2em}Topology: \V{EX\_TOPOLOGICAL\_DESCRIPTION}\\
    \hspace*{1.2em}GT label: \V{EX\_GT\_LABEL}; GNN pred: \V{EX\_GNN\_PRED}; class probs: \V{EX\_CLASS\_PROBS}\\
    (Repeat for each example.)\\[0.6ex]

    [Rewrite instructions]\\
    Rewrite the target description using only the provided inputs.\\
    Output one natural-language paragraph (no bullet points), $<200$ words.
  \end{tcolorbox}

  \caption{Unified prompt used to refine node text descriptions. Only placeholders (e.g., \V{GRAPH\_TYPE}) are filled at runtime; the prediction and example blocks are optional.}
  \label{fig:prompt-templates}
\end{figure*}

\end{document}